\documentclass[runningheads]{llncs}
\pdfoutput=1
\usepackage[T1]{fontenc}
\usepackage[sorting=none, style=numeric,backend=biber,maxnames=99]{biblatex}
\usepackage[T1]{fontenc}
\usepackage{amsmath}
\usepackage{graphicx}
\usepackage{caption}
\usepackage{float} 
\usepackage{subfigure}
\usepackage{subcaption}
\usepackage[usenames,dvipsnames]{xcolor}
\usepackage[final]{pdfpages}
\usepackage{multirow}
\usepackage{graphicx}
\usepackage{algorithmic}
\usepackage{algorithm}
\usepackage{marvosym}

\newcommand{\notem}[1]{{\color{green} #1 [MB].}}
\newcommand{\mike}[1]{{\color{blue} #1 [MB].}}
\newcommand{\JH}[1]{{\color{red} #1 [JH].}}
\AtBeginDocument{%
  }
\usepackage{graphicx}
\begin{document}
\title{\textit{GIG}: \underline{G}raph Data \underline{I}mputation With \underline{G}raph Differential Dependencies}
%
%
\author{Jiang Hua\inst{1} \and Michael Bewong\inst{2,3} \and Selasi Kwashie\inst{2} \and MD Geaur Rahman \inst{3} \and Junwei Hu\inst{1} \and Xi Guo\inst{1}\textsuperscript{\Letter} \and Zaiwen Feng\inst{1}\textsuperscript{\Letter}}

%
%
\institute{College of Informatics, Huazhong Agricultural University, Wuhan, Hubei, China 
\email\{zaiwen.feng,xguo\}@mail.hzau.edu.cn  \and
AI and Cyber Futures Institute, Charles Sturt University, NSW,
Australia     \and
School of Computing, Mathematics and Engineering, Charles Sturt University, NSW, Australia}
\maketitle              
\begin{abstract}
Data imputation addresses the challenge of imputing missing values in database instances, ensuring consistency with the overall semantics of the dataset. 
Although several heuristics which rely on statistical methods, and ad-hoc rules have been proposed. These do not generalise well and often lack data context. Consequently, they also lack explainability. The existing techniques also mostly focus on the relational data context making them unsuitable for wider application contexts such as in graph data.
In this paper, we propose a graph data imputation approach called \emph{GIG} which relies on graph differential dependencies (GDDs). \emph{GIG}, learns the GDDs from a given knowledge graph, and uses these rules to train a transformer model which then predicts the value of missing data within the graph.  
By leveraging GDDs, \emph{GIG} incoporates semantic knowledge into the data imputation process making it more reliable and explainable.  
Experimental results on seven real-world datasets highlight GIG's effectiveness compared to existing state-of-the-art approaches.

\end{abstract}
\section{Introduction}

In the era of big data, addressing the challenge of missing value imputation has become critical \cite{batista2003analysis}. 
In general, datasets with incomplete information pose challenges for deriving reliable knowledge \cite{montesdeoca2019first}. Consequently, considerable efforts have been directed towards the data imputation problem, viewed as a fundamental task in data cleaning \cite{chu2016data}. The complexity of identifying the optimal values for imputing missing data arises from the necessity to evaluate all potential combinations within the value distribution. Numerous approaches in the literature operate under the premise that a missing value can be imputed with another value from the same population, with the overarching goal of maintaining the overall integrity of the data \cite{donders2006gentle}. Furthermore, heuristics have been extensively utilized to judiciously choose values for imputation, aiming to enhance the efficiency of the process without compromising the accuracy of the imputed values \cite{cihan2019new}. However, most of these techniques focus on relational data scenarios which is not bewildered with the same challenges pertaining to graph data. Thus, the task of addressing data imputation in graph data is a non-trivial task. 

This paper presents GIG, a data imputation algorithm for completing missing values in knowledge graphs, based on a Graph Differential Dependency-guided (GDD-guided)  \emph{transformer} model. GDDs go beyond Graph Entity Dependencies (GEDs) by integrating distance and matching functions in lieu of equality functions. 
In general, GDDs exhibit better compatibility with data and are more suitable for discovering and handling various types of errors, such as duplicates, outliers, and constraint violations\cite{ilyas2015trends}. Hence, GDDs have the potential to be employed for identifying appropriate candidate values to replace missing ones in the data imputation process. 
Our proposed technique GIG leverages GDDs to identify instances that satisfy GDD rules to train a transformer model; formulates missing values in the form of rules and leverages the trained transformer to complete the rule; and relies on the GDD rules to validate the  semantic consistency of the missing value candidates.

The paper is structured as follows: Section 2 reviews the literature on data imputation approaches. Section 3 introduces preliminary notions on GDDs. Section 4 delineates the data imputation problem. The GIG algorithm is detailed in Section 5, and an experimental evaluation measuring its effectiveness is presented in Section 6. Finally, conclusions and directions for further research are discussed in Section 7.

\section{RELATED WORK}
In the last decade, various solutions have been proposed for data imputation, including the application of linear regression methods. Specifically, the linear regression model in \cite{zhang2019learning} addresses imputation of numerical missing values to tackle the data sparsity problem, where the number of complete tuples may be insufficient for precise imputation. 
Further, \emph{REMIAN} proposed in~\cite{ma2020remian} employs a multivariate regression model (MRL) to handle real-time missing value imputation in error-prone environments, dynamically adapting parameters and incrementally updating them with new data. 
The work in \cite{song2018enriching} utilizes regression models differently for imputing missing values. Rather than directly inferring missing values, the authors suggest predicting distances between missing and complete values and then imputing values based on these distances.  Other techniques such as \cite{marcelino2022missing} use regression models to analyse missing values.


In~\cite{samad2022missing}, Samad proposes a hybrid framework that combines ensemble learning and deep neural networks (DNN) with the Multiple Imputation using Chained Equations (MICE) approach to improve imputation and classification accuracy for missing values in tabular data \cite{samad2022missing}. They introduce cluster labels from training data 
to enhance imputation accuracy and minimize variability in missing value estimates. The paper demonstrates superior performance of their methods compared to baseline MICE and other imputation algorithms, particularly for high missingness and non-random missing types. 

The Holoclean framework, introduced in \cite{rekatsinas2017holoclean} employs machine learning techniques to rectify errors in structured datasets. It addresses inconsistencies like duplicates, incorrect entries, and missing values by automatically generating a probabilistic model. This model extrapolates features representing dataset uncertainty, which are used to characterize a probabilistic graphical model. Statistical learning and probabilistic inference are then applied to repair errors. Additionally, data entry consistency can be maintained by defining integrity constraints. The Holoclean framework has also been shown to be useful in multi-source heterogeneous data~\cite{cui2022holocleanx}.

 The techniques discussed so far 
 aim to address the data imputation problem by using models that infer relationships among data in a supervised or unsupervised manner. These approaches are notably fast, although with the exception of Holoclean, they are restricted to numerical datasets by nature. In contrast, RENUVER proposed in ~\cite{breve2022renuver} is suited for numerical, textual, and categorical data, treating each missing value according to the data domain it belongs to. Although the methodology employed in Holoclean can permit the imputation of textual and categorical values with the use of metadata, specifically \emph{denial constraints}, 
 the role of metadata in the imputation process is limited, and  primarily serves as integrity constraints that imputed values must satisfy. 
 In contrast, metadata can be more extensively leveraged to generate candidate values, as demonstrated in RENUVER with relational functional dependencies (RFDs). 

Indeed, RFDs have been explored in data imputation. Bohannon et al. introduced Conditional Functional Dependencies (CFDs), a specific type of RFDs, capable of capturing overall data correctness \cite{wang2018imputing}. While CFDs capture data semantics and are suitable for cleaning, the authors didn't propose a dedicated imputation approach. They defined SQL queries for CFD violation detection, potentially used for checking imputed dataset integrity. However, determining CFDs involves an expensive, manual process \cite{fan2010discovering, al2022data}. Another imputation algorithm using RFDs is by Song et al. \cite{song2011differential}. They used differential dependencies (DDs) \cite{song2023integrity}, a type of RFDs \cite{bleifuss2017efficient}, employing similarity rules for value tolerance. To maximize imputed missing values, the authors presented four algorithms, including integer linear programming, its approximation, a randomized method, and Derand, a derandomized version considered optimal for deterministic approximation of multiple missing values.

\section{PRELIMINARIES}
In this section we introduce the concept of GDDs and show how GDDs can be used for graph data imputation.

A GDD $\sigma$ is a pair ( $Q[\bar{z}]$ , $\Phi_X \rightarrow \Phi_Y$), where: $Q[\bar{z}]$ is a graph pattern called the scope, $\Phi_X \rightarrow \Phi_Y$ is called the dependency, $\Phi_X$ and $\Phi_Y$ are two (possibly empty) sets of distance constraints on the pattern variables $\bar{z}$.  A \textit{distance constraint} in $\Phi_X$ and $\Phi_Y$ on $\bar{z}$ is one of the following~\cite{kwashie2019certus}:

\begin{tabular}{r r}
   \(d_A(x.A,c) \leq t_A;\) & \(d_{A_iA_j}(x.A_i,x{'}.A_j) \leq t_{A_iA_j};\)\\
   
   \(d_\equiv(x.eid,C_e) = 0;\)& \(d_\equiv(x.eid,x{'}.eid) = 0;\)\\
   
   \(d_\equiv(x.rela,C_r) = 0;\)& \(d_\equiv(x.rela,x{'}.rela) = 0;\)\\\\
\end{tabular}

where \(x,x' \in \bar{z},\ A, A_{i}, A_{j}\) are attributes in \textbf{A}, c is a value of \(A,\ d_{A_iA_j} (x.A_i,x{'}.A_j)\) (or \( d_{A_i}(x,x{'})\) if \(A_i = A_j)\) is a user specified distance function for values of \((A_i, A_j),\ t_{A_iA_j}\) is a threshold for \((A_i, A_j)\), \(d_\equiv(\cdot,\cdot)\) are functions on \textbf{eid} and relations and they return 0 or 1. \(d_\equiv(x.eid,C_e) = 0\) if the eid value of $x$ is $C_e$, \(d_\equiv(x.eid,x{'}.eid) = 0\) if both $x$ and \(x{'}\) have the same eid value, \(d_\equiv(x.rela,C_r) = 0\) if $x$ has a relation named rela and ended with the profile/node $C_r$, \(d _\equiv(x.rela,x{'}.rela) = 0\) if both \(x\) and \(x{'}\) have the relation named rela and ended with the same profile/node.


The user-specified distance function \(d_{A_iA_j}(x.A_i, x^{'}.A_j)\) depends on the types of $A_i$ and $A_j$. It can take the form of an arithmetic operation involving interval values, an edit distance calculation for string values, or the distance computation between two categorical values within a taxonomy, among other possibilities. These functions accommodate the wildcard value `*' to represent any domain, and in such cases, they return a 0 distance.

We call $\Phi_X$ and $\Phi_Y$ the LHS and the RHS functions of the dependency respectively. In this work we rely on the GDDMiner proposed in ~\cite{zhang2023discovering} to mine the GDD rules from graph data.

\begin{figure*}[htbp]
    \centering    \includegraphics[width=1\textwidth,height=0.55\textwidth]{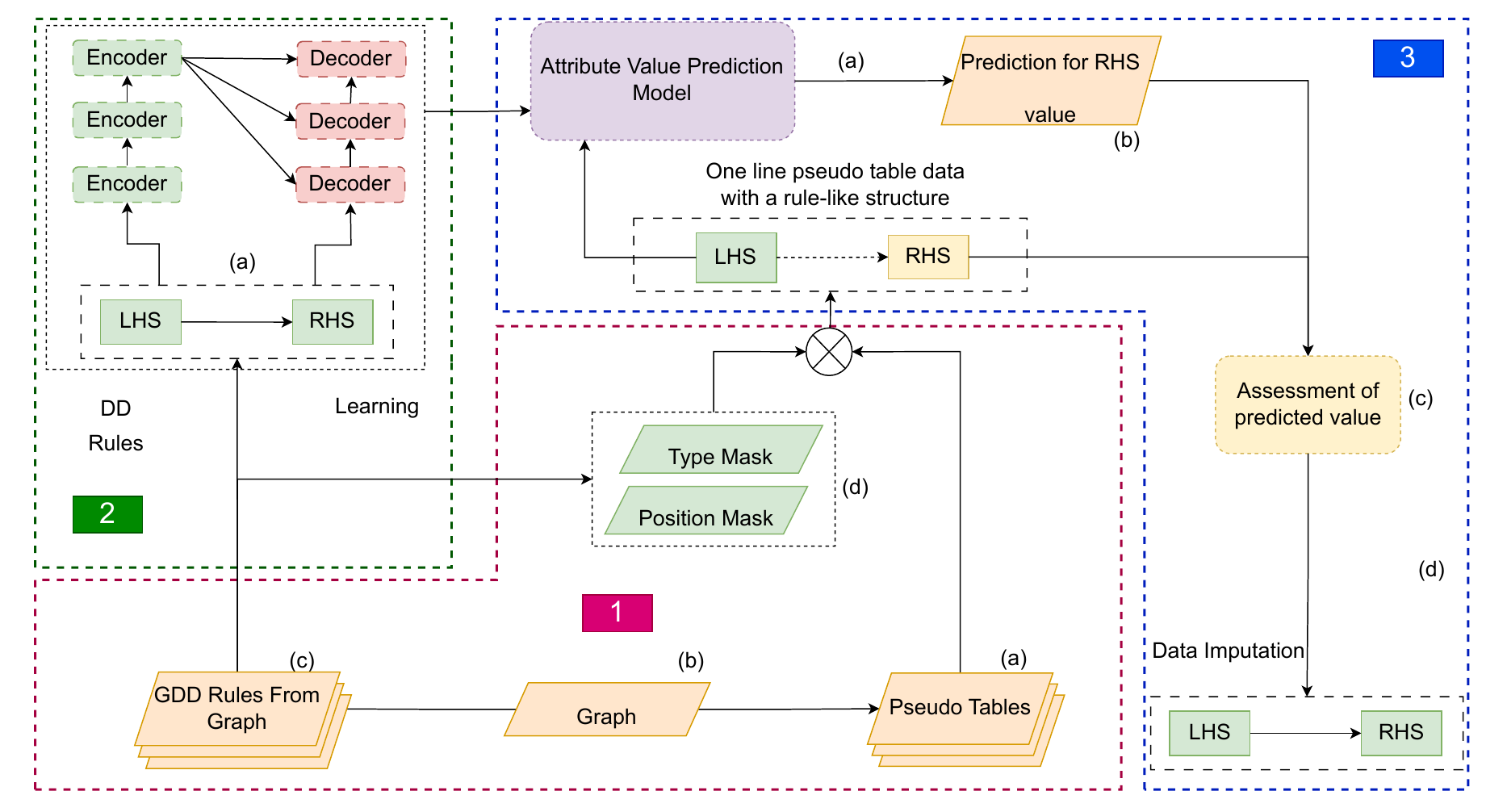}
    \caption{\textit{GIG} Framework}
    \label{fig:framework}
\end{figure*}

\section{\textit{GIG} ALGORITHM}
GIG relies on a transformer architecture and graph differential dependencies (GDDs) for data imputation. Given a knowledge graph $G$, GIG proceeds in three steps namely (1) rule mining and selection; (2) transformer training; and (3) missing value imputation. These steps are summarised in Figure \ref{fig:framework} and explained as follows. 

\subsection{Rule Mining and Selection}

The GDD mining algorithm proposed in \cite{zhang2023discovering} is employed to extract GDD rules from the knowledge graph $G$ as shown in  Figure \ref{fig:framework}, Part \fcolorbox{black}{RubineRed}{\textcolor{white}{1}}.
For example \(d(y.name, \hat{y}.name) \leq 1 \rightarrow d(y.genre, \hat{y}.genre) = 0\) is  one such rule that implies any two entities $y$ and $\hat{y}$ with a similar name will belong to the same genre. 

GDD rules are composed of a left-hand side (LHS) and a right-hand side (RHS). The LHS serves as the input to the transformer encoder, while the RHS serves as the input to the transformer decoder (\emph{cf.} Figure \ref{fig:framework}, Part \fcolorbox{black}{OliveGreen}{\textcolor{white}{2}} (a)). 
When dealing with multiple GDD rules, where two or more rules exhibit distinct left-hand sides (LHS) but share the same right-hand side (RHS), we opt to consolidate the LHS to enhance the predictive outcomes of the transformer model. 
Similarly, in cases where multiple rules feature the same LHS but different RHS, we amalgamate the RHS. For example, a rule set 
\(X \rightarrow Y_1\), \(X \rightarrow Y_2\), \(X \rightarrow Y_3\), will be merged to \(X \rightarrow Y_1, Y_2, Y_3\). 
This adaptation is implemented to consolidate the rules and streamline the learning process of the transformer model.

In the case of GDD rules, we employ masking vectors for the storage of attribute and type details. Both the left-hand side (LHS) and right-hand side (RHS) of GDD rules encompass attribute information, and the utilization of masking vectors for this purpose is both succinct and efficient. 
For instance, consider a tuple h1 from Table \ref{table:pseduo}. Let us assume we have a GDD rule \(r: x.name \rightarrow y.genre\), in this case, we can construct a mask for the attributes, represented by (0, 1, 0, 0, 1, 0, 0, 0, 0, 0, 0), where a value of 1 indicates that the attribute is utilized. This approach enables us to preserve a substantial amount of information related to GDD rules. 
\begin{table*}[!htbp]
    \centering
    \caption{pseudo-relational table}
    \resizebox{\linewidth}{!}{
    \begin{tabular}{|c|c|c|c|c|c|c|c|c|c|c|c|}
    \hline
        Pattern variables & \multicolumn{2}{c|}{x}  & \multicolumn{5}{c|}{y} & \multicolumn{4}{c|}{y'} \\ \hline
        Attributes / Matches & id & Name (A1) & id & Name (A2) & Genre (A3) & Year (A4) & Price (A5) & id & Name (A2) & Genre (A3) & Year (A4) \\ \hline
        h1 & 1 & GL & 4 & AF9 & Racing & 2018 & £50 & 5 & AF11 & Racing & 2018 \\ \hline
        h2 & 3 & EA & 7 & ? & Soccer & 2019 & £55 & 8 & F20 & Soccer & 2019 \\ \hline
        h3 & 3 & EA & 7 & F20 & Soccer & 2019 & £55 & 9 & F21 & Soccer & 2020 \\ \hline
        h4 & 3 & EA & 8 & F20 & Soccer & 2019 & £60 & 9 & F21 & Soccer & 2020 \\ \hline
    \end{tabular}}
    \label{table:pseduo}
\end{table*}


\subsection{Transformer Training}

The transformer consists of an encoder and a decoder (\emph{cf.} Figure \ref{fig:framework}, Part \fcolorbox{black}{OliveGreen}{\textcolor{white}{2}}), where the LHS of the rules serves as the input to the encoder, and the RHS as the input to the decoder. Thus, for a GDD rule, \((Q, X\rightarrow Y)\), the transformer's encoder learns the features of $X$ and the decoder learns the features of $Y$. In this way, the transformer learns the relationship between the LHS and RHS of the rules. 

\textit{Encoder.} Given a rule \((Q, X\rightarrow Y)\), $X$ contains one or more distinct attribute values, which are all considered as a single unit during the training phase. For example, the LHS $X := d(y.name, y'.name) \leq 1$, will be treated as a literal $\{y.name, y'.name, \leq 1\}$. 
By treating $(y.name, y'.name)$ and $\leq 1$ as unified literal within the encoding framework, we enable the transformer to infer their interconnectedness. We have observed, empirically, that this approach is more robust against the potential of variations within the attribute values which can deteriorate model performance.


\textit{Decoder.} The decoder uses a similar approach to the encoder. When handling the RHS $Y := d(y.gender, y'.gender) = 0$, by treating $\{y.gender, y'.gender, =0\}$  as unified literal.


\textit{Training.} Once the inputs to the encoder (LHS) and decoder (RHS) are derived from the rules, the transfomer model learns GDD rules by minimising the loss function \emph{i.e.} Kullback-Leibler Divergence  (KLDivLoss). KLDivLoss measures the divergence between the predicted distribution and the target distribution. Lower values of the KLDivLoss indicate better alignment between input and output, which is desirable during training. The final result after training serves as the predictive model.

\subsection{Imputing missing value}
\textit{Imputing missing value.}
Missing values can also be expressed as rules for example $X \rightarrow \_ $ implies the LHS values $X$ are known and the values of $Y$ are to be determined. GIG uses the LHS in the trained transformer predictive model to predict an RHS value for imputing $Y$, facilitating the identification of all viable candidates for the imputation.  

Let us consider a scenario where we are interested in predicting the y.name value of record h2 in the pseudo-relational table Table~\ref{table:pseduo}. The first step is to identify all GDD rules that has y.name in its RHS. That is any rule that has a `1' in its positional mask at index 3 is considered. For example, the rule $x.name \rightarrow \ y.name $ which has the positional mask (0, 1, 0, 1, 0, 0, 0, 0, 0, 0, 0) will be considered since it has a value `1' in the index position 3. Intuitively, the rule means any two nodes of type $x$ that share the same name, will also share the same name on a node type $y$.

Next, based on the rules identified to be relevant, we can identify the LHS value which should serve as input to the transformer model. In this case EA becomes the input since according to the rule, the LHS should consists of x.name, and in h2 x.name is EA. The transformer then makes a prediction of what the RHS value should be. The predicted value can either be  \emph{semantically consistent} or \emph{inconsistent}.  For instance, consider the scenario where the transformer predicts a value of "2020". In this scenario, when this value is imputed, it violates the rule $x.name \rightarrow \ y.name $ since "2020" is not a \textit{name} attribute but rather a \textit{year} attribute. Thus this value is discarded. However, if the transformer model predicts the value "F20", this value can be accepted as an imputed value since it does not violate the rule $x.name \rightarrow \ y.name $. It is possible that there can be several matching rules and several predicted candidates, for simplicity in this work, we only rely on the top 1 rule and predicted candidates for imputation. 

The GIG algorithm is summarised in Algorithm~\ref{algorithm}~\footnote{All source codes and datasets used in this paper will be made publicly available upon acceptance of the paper}.

\begin{algorithm}
\caption{GIG} 
\label{algorithm}

\begin{algorithmic}[1]
  \ENSURE{ Graph differential dependencies $GDDs$, pseudo table $T$ and knowledge graph $G$ with missing values}
  \REQUIRE{ Imputed knowledge graph $G'$}  

  \STATE Construct encoder input $W_l$, decoder output $W_r$, and mask $B$ from $GDDs$ 
  \FOR{each epoch}
        \FOR{each batch \(\in W_l \) and $W_r$}
            \STATE Get encoder input and decoder output
            \STATE Perform forward pass through encoder and decoder; Compute loss
            \STATE Perform backward pass and parameter update
        \ENDFOR
    \ENDFOR
  \STATE $\mathcal{M}$ $\xleftarrow{save}$ transformer.model
  \FOR{\(t \in T\)}
        \IF{$t$ has missing value}
        \STATE {$M \leftarrow $ index position(s) of missing value(s)}
        \FOR{\(b \in B\)}
            \IF{$b.M \neq 0$  }
            \STATE Get rule(s) $X \rightarrow Y$ from $b$ such that $Y$ contains $b.M$
            \STATE {$I \leftarrow $ index position(s) of $X.attributes$}
            \RETURN
            \ENDIF
         \ENDFOR   
    \STATE $t'.M$ $\xleftarrow{predicted}$ $\mathcal{M}(\{t.I\})$ 
        \STATE Impute $G'$ with prediction $t'.M$ 
        \ENDIF
        
  \ENDFOR
  \RETURN $G'$
\end{algorithmic} 

\end{algorithm}
The worst-case time complexity of GIG is \(O(n \cdot m \cdot \Sigma)\), where \(n\) represents the number of GDD rules, \(m\) represents the number of rows in the pseudo-relational table, and \(\Sigma\) represents the time taken for one run of the transformer. 
To elaborate further, 
each row in the pseudo-relational table is compared to each rule in the worst case when finding matching rules for a missing value, the   time complexity is \(O(n \cdot m)\).
Next, the time complexity of running the transformer model on a single output from the GDD rule and pseudo relational table comparison is \(\Sigma\). 
Combining both steps, the overall worst-case time complexity of GIG is \(O(n \cdot m \cdot \Sigma)\). 

\section{Experiment}

\subsection{Experimental Setting}
All experiments were performed on a single server, Intel(R) Xeon(R) Silver 4210R CPU @ 2.40GHz, 32GB RAM. We also adopt \emph{transformer} architecture following the work in~\cite{vaswani2017attention}.

\subsubsection{Datasets.}
In this paper, we use seven real world datasets summarised in Table~\ref{tab_datasets} for evaluation. 
Namely (1) \emph{Restaurant}  which includes six attributes, including the name and address of the restaurant; (2) \emph{Customer\_shopping\_data (CSD)} which contains six attributes, including the customer's age and the product price; (3) \emph{Adult} which has 11 attributes, including an adult's occupation and nationality; (4) \emph{Ncvoter} which has six attributes, including the patient's age and registration date; (5) \emph{Entity Resolution}, a dataset derived from a virtual online video streaming platform; (6) \emph{Women’s World Cup} which contains five node types representing the {\emph{person}}, {\emph{team}}, {\emph{squad}}, {\emph{tournament}} and {\emph{match}} from all the World Cups between 1991 and 2019; (7)
\emph{Graph Data Science} dataset which reflects the connections between different airports around the world. 

For all datasets, we vary the percentage of missing values in the range  [1\%, 5\%] on each considered dataset, with random selection of the data values.
We note that, \emph{Restaurant}, \emph{Customer\_shopping\_data}, \emph{Adult} and \emph{Ncvoter} summarised in Table \ref{table:pseduo}, are relational data that have also been used in benchmark techniques \cite{song2018enriching, rekatsinas2017holoclean}. For these datasets, we first convert them into knowledge graphs, from which we extract GDD rules and generate pseudo-relational tables \cite{zhang2023discovering}. 

\begin{table*}[htbp]
    \centering    
    \caption{Summary of real-world datasets}\label{tab_datasets}
    \begin{tabular}{|l|c|c|c c c c c|}
    \hline
        \multirow{2}{*}{\textbf{Dataset}} &  \textbf{\# Attributes} &  \textbf{\# Tuples/Nodes} &  \multicolumn{5}{c|}{\textbf{\# missing values} }\\ 
        ~ & ~ & ~ &  \textbf{[1\%]} &  \textbf{[2\%]} &  \textbf{[3\%]} &  \textbf{[4\%]} &  \textbf{[5\%]} \\ \hline
        Restaurant & 6 & 864 & 52 & 104 & 155 & 206 & 259 \\ 
        Customer\_shopping\_data & 6 & 500 & 45 & 90 & 135 & 180 & 225 \\ 
        Adult & 11 & 500 & 55 & 110 & 165 & 220 & 275 \\ 
        Ncvoter & 7 & 500 & 35 & 70 & 105 & 140 & 175 \\ 
        Entity Resolution & 13 & 676 & 88 & 174 &261 & 348 & 435 \\
        Graph data Science & 7 & 519 & 36 & 72 & 108 & 144 & 180 \\
        WomensWorldCup2019 & 13 & 514 & 66 & 132 & 198 & 254 & 330 \\\hline
    \end{tabular}
\end{table*}

\subsubsection{Evaluation Baselines}
In this work, we rely on Holoclean~\cite{cihan2019new}  and DERAND \cite{song2018enriching} as baselines for comparison. 
DERAND employs tolerance-based similarity rules instead of strict editing rules to discard invalid candidates identified by similarity neighbors. Holoclean is a Holistic machine-learning solution for data imputation~\cite{cihan2019new}.

\subsubsection{Evaluation metrics.} The effectiveness of the data imputation approaches was evaluated using three different metrics: precision, recall, and F1-measure, which is consistent with the existing literature \cite{breve2022renuver}.

In this context, precision quantifies how many missing values are correctly imputed relative to the total number of imputed missing values. This parameter serves as a reliability score, indicating how well the algorithm performs when deciding whether to impute with an uncertain value or leave the missing value. That is, if "true" represents the correctly imputed missing values and "imputed" represents all the imputed missing values, then the precision is calculated as follows: $\text{precision} = \frac{| \text{true} \cap \text{imputed} |}{| \text{imputed} |} $.

Recall represents the fraction of correctly imputed missing values. That is, if "missing" represents the missing values in the dataset and "true" represents the correctly imputed missing values at the end of the imputation process, then the recall is calculated as follows: $\text{recall} = \frac{| \text{true} \cap \text{missing} |}{| \text{missing} |}$.

The F1-measure is computed by combining precision and recall according to the following formula: $\text{F1-measure} = \frac{2 \times \text{precision} \times \text{recall}}{\text{precision} + \text{recall}}$.

\begin{figure*}[htb]
  \centering
\includegraphics[width=0.9\textwidth,height=0.8\textwidth]{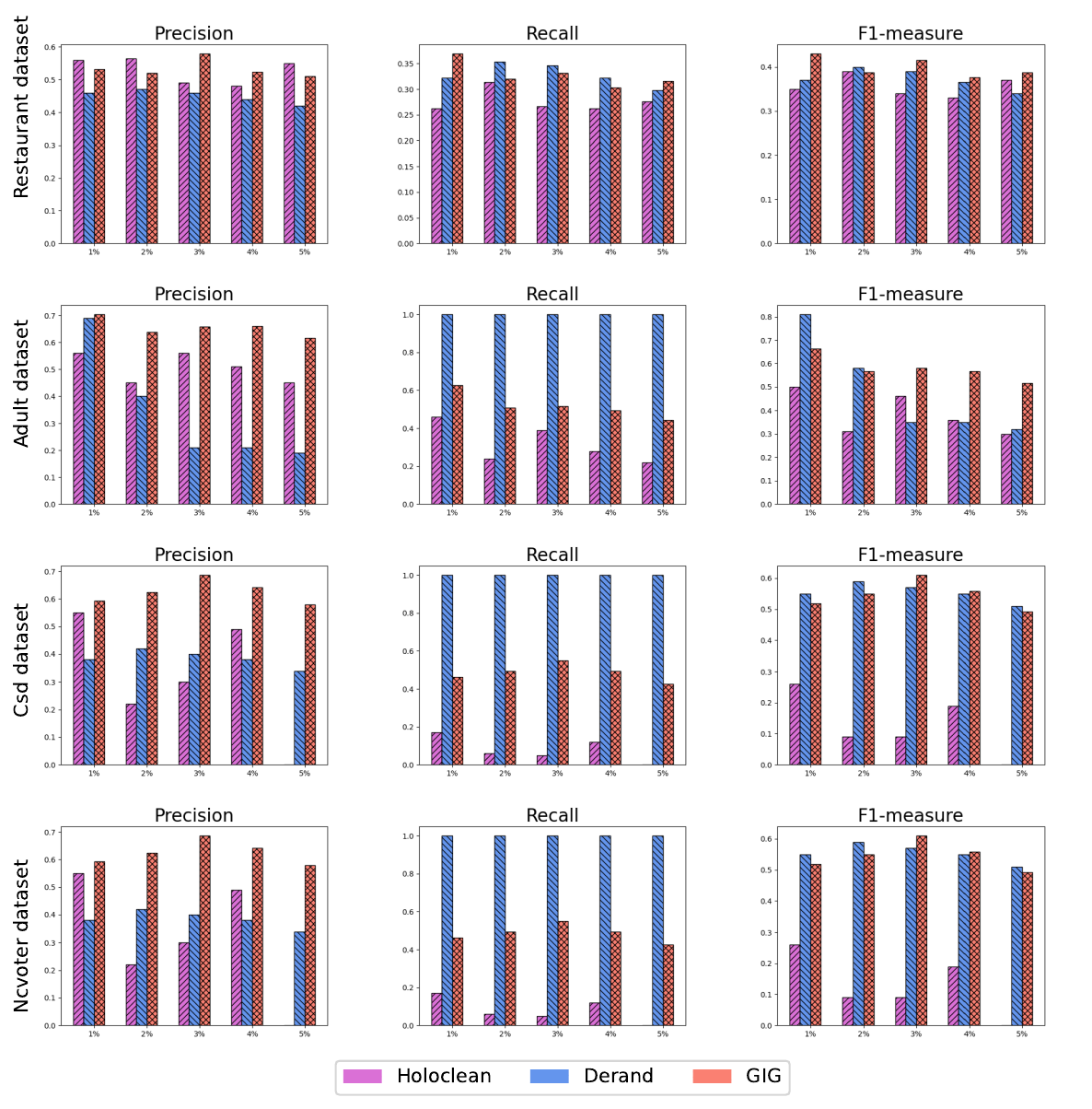}
  \caption{ Comparison of GIG with Derand and Holoclean}
  \label{fig:compare}
\end{figure*}

\subsection{Effectiveness on Relational Data}
In this section we compared our proposed method GIG with known relational data based techniques 
such as Derand \cite{song2018enriching}  and Holoclean \cite{rekatsinas2017holoclean}. 
Figure~\ref{fig:compare} are the results. 
In Figure~\ref{fig:compare} we observe that GIG performs competitively with Holoclean and Derand. In particular, across all datasets, GIG has a higher precision rate than other techniques due to the rule guided transformer approach. We do notice, however that with the exception of \emph{Restaurant} dataset, the rule guided approach also leads to a comparatively lower recall on \emph{Adult},\emph{CSD} and \emph{Ncvoter} datasets. Yet still, GIG is highly competitive and in most cases has a better performance in F1 across all the datasets.

\subsection{Effectiveness on Graph Datasets:}
\begin{figure*}[htbp]
  \centering
\includegraphics[width=1\textwidth]{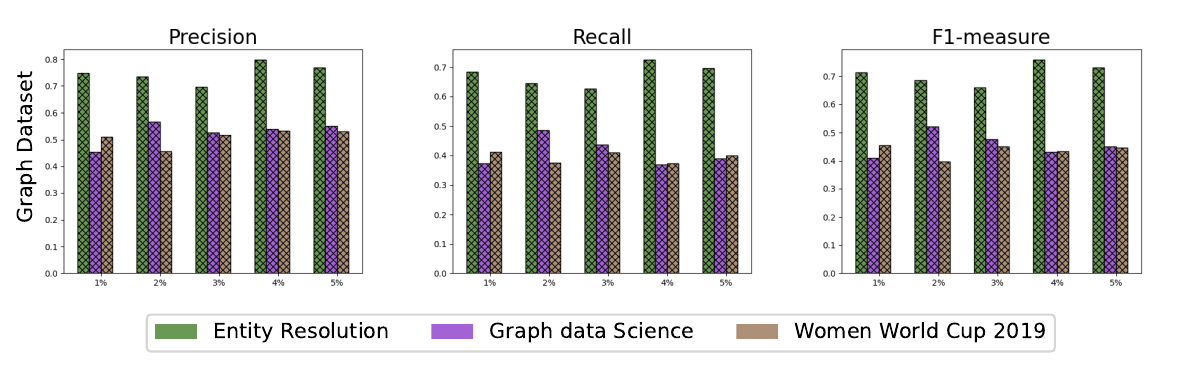}
  \caption{ Performance of GIG on Graph Datasets}
  \label{fig:graph}
\end{figure*}

We assess the effectiveness of GIG for imputing missing values in graph data. We evaluate GIG's performance on the three graph datasets \emph{Entity Resolution}, \emph{Graph Data Science}, and \emph{WomensWorldCup2019}.  Figure~\ref{fig:graph} is the result of that experiment. In the figure, we observe that the results for \emph{Entity Resolution} were the best, as its data distribution yields more exemplar points that enables the GDD-guided transformer model to represent the data more effectively. This is in contrast with the other two datasets. 

\section{Conclusion}
In this paper, we introduced GIG, a data imputation algorithm that leverages graph differential dependencies for missing value imputation. GIG relies on a rule-guided transformer architecture and has three main steps involving GDD rule mining, transformer training and missing value imputation. 
Evaluation results from seven datasets and in comparison with existing state-of-the-art demonstrate the competitiveness and suitability of GIG for both relational and graph data imputation.  
In our future work, we aim to enhance GIG by considering different types of rules such as approximate graph entity rules which may improve the overall recall of GIG.  

\section{Acknowledgement.}
This research project was supported in part by National Key Research and Development Program of China under Grant 2023YFF1000100, and in part by the Fundamental Research Funds for the Chinese Central Universities under Grant 2662023XXPY004.

\section*{REFERENCES}

\printbibliography[heading=none]

@article{kwashie2019certus,
  title={Certus: An effective entity resolution approach with graph differential dependencies (GDDs)},
  author={Kwashie, Selasi and Liu, Lin and Liu, Jixue and Stumptner, Markus and Li, Jiuyong and Yang, Lujing},
  journal={Proceedings of the VLDB Endowment},
  volume={12},
  number={6},
  pages={653--666},
  year={2019},
  publisher={VLDB Endowment}
}

@inproceedings{breve2022renuver,
  title={RENUVER: A Missing Value Imputation Algorithm based on Relaxed Functional Dependencies.},
  author={Breve, Bernardo and Caruccio, Loredana and Deufemia, Vincenzo and Polese, Giuseppe},
  booktitle={EDBT},
  pages={1--52},
  year={2022}
}

@inproceedings{zhang2023discovering,
  title={Discovering Graph Differential Dependencies},
  author={Zhang, Yidi and Kwashie, Selasi and Bewong, Michael and Hu, Junwei and Mahboubi, Arash and Guo, Xi and Feng, Zaiwen},
  booktitle={Australasian Database Conference},
  pages={259--272},
  year={2023},
  organization={Springer}
}

@article{batista2003analysis,
  title={An analysis of four missing data treatment methods for supervised learning},
  author={Batista, Gustavo EAPA and Monard, Maria Carolina},
  journal={Applied artificial intelligence},
  volume={17},
  number={5-6},
  pages={519--533},
  year={2003},
  publisher={Taylor \& Francis}
}

@inproceedings{chu2016data,
  title={Data cleaning: Overview and emerging challenges},
  author={Chu, Xu and Ilyas, Ihab F and Krishnan, Sanjay and Wang, Jiannan},
  booktitle={Proceedings of the 2016 international conference on management of data},
  pages={2201--2206},
  year={2016}
}

@article{cihan2019new,
  title={A new heuristic approach for treating missing value: ABCIMP},
  author={Cihan, Pinar and Ozger, Zeynep Banu},
  journal={Elektronika ir Elektrotechnika},
  volume={25},
  number={6},
  pages={48--54},
  year={2019}
}

@article{donders2006gentle,
  title={A gentle introduction to imputation of missing values},
  author={Donders, A Rogier T and Van Der Heijden, Geert JMG and Stijnen, Theo and Moons, Karel GM},
  journal={Journal of clinical epidemiology},
  volume={59},
  number={10},
  pages={1087--1091},
  year={2006},
  publisher={Elsevier}
}

@inproceedings{montesdeoca2019first,
  title={A First Approach on Big Data Missing Values Imputation.},
  author={Montesdeoca, Besay and Luengo, Juli{\'a}n and Maillo, Jes{\'u}s and Garc{\'\i}a-Gil, Diego and Garc{\'\i}a, Salvador and Herrera, Francisco},
  booktitle={IoTBDS},
  pages={315--323},
  year={2019}
}

@article{ilyas2015trends,
  title={Trends in cleaning relational data: Consistency and deduplication},
  author={Ilyas, Ihab F and Chu, Xu},
  journal={Foundations and Trends{\textregistered} in Databases},
  volume={5},
  number={4},
  pages={281--393},
  year={2015},
  publisher={Now Publishers, Inc.}
}

@article{rekatsinas2017holoclean,
  title={Holoclean: Holistic data repairs with probabilistic inference},
  author={Rekatsinas, Theodoros and Chu, Xu and Ilyas, Ihab F and R{\'e}, Christopher},
  journal={arXiv preprint arXiv:1702.00820},
  year={2017}
}

@article{song2018enriching,
  title={Enriching data imputation under similarity rule constraints},
  author={Song, Shaoxu and Sun, Yu and Zhang, Aoqian and Chen, Lei and Wang, Jianmin},
  journal={IEEE transactions on knowledge and data engineering},
  volume={32},
  number={2},
  pages={275--287},
  year={2018},
  publisher={IEEE}
}

@article{ma2020remian,
  title={REMIAN: Real-time and error-tolerant missing value imputation},
  author={Ma, Qian and Gu, Yu and Lee, Wang-Chien and Yu, Ge and Liu, Hongbo and Wu, Xindong},
  journal={ACM Transactions on Knowledge Discovery from Data (TKDD)},
  volume={14},
  number={6},
  pages={1--38},
  year={2020},
  publisher={ACM New York, NY, USA}
}

@inproceedings{zhang2019learning,
  title={Learning individual models for imputation},
  author={Zhang, Aoqian and Song, Shaoxu and Sun, Yu and Wang, Jianmin},
  booktitle={2019 IEEE 35th International Conference on Data Engineering (ICDE)},
  pages={160--171},
  year={2019},
  organization={IEEE}
}

@inproceedings{wang2018imputing,
  title={Imputing structured missing values in spatial data with clustered adversarial matrix factorization},
  author={Wang, Qi and Tan, Pang-Ning and Zhou, Jiayu},
  booktitle={2018 IEEE International Conference on Data Mining (ICDM)},
  pages={1284--1289},
  year={2018},
  organization={IEEE}
}

@article{fan2010discovering,
  title={Discovering conditional functional dependencies},
  author={Fan, Wenfei and Geerts, Floris and Li, Jianzhong and Xiong, Ming},
  journal={IEEE Transactions on Knowledge and Data Engineering},
  volume={23},
  number={5},
  pages={683--698},
  year={2010},
  publisher={IEEE}
}

@article{song2011differential,
  title={Differential dependencies: Reasoning and discovery},
  author={Song, Shaoxu and Chen, Lei},
  journal={ACM Transactions on Database Systems (TODS)},
  volume={36},
  number={3},
  pages={1--41},
  year={2011},
  publisher={ACM New York, NY, USA}
}

@article{bleifuss2017efficient,
  title={Efficient denial constraint discovery with hydra},
  author={Bleifu{\ss}, Tobias and Kruse, Sebastian and Naumann, Felix},
  journal={Proceedings of the VLDB Endowment},
  volume={11},
  number={3},
  pages={311--323},
  year={2017},
  publisher={VLDB Endowment}
}

@article{marcelino2022missing,
  title={Missing data analysis in regression},
  author={Marcelino, Carolina Gil and Leite, Gabriel MC and Celes, P and Pedreira, Carlos Eduardo},
  journal={Applied Artificial Intelligence},
  volume={36},
  number={1},
  pages={2032925},
  year={2022},
  publisher={Taylor \& Francis}
}

@article{samad2022missing,
  title={Missing value estimation using clustering and deep learning within multiple imputation framework},
  author={Samad, Manar D and Abrar, Sakib and Diawara, Norou},
  journal={Knowledge-based systems},
  volume={249},
  pages={108968},
  year={2022},
  publisher={Elsevier}
}

@inproceedings{cui2022holocleanx,
  title={HoloCleanX: A Multi-source Heterogeneous Data Cleaning Solution Based on Lakehouse},
  author={Cui, Qin and Zheng, Wenkui and Hou, Wei and Sheng, Ming and Ren, Peng and Chang, Wang and Li, XiangYang},
  booktitle={International Conference on Health Information Science},
  pages={165--176},
  year={2022},
  organization={Springer}
}

@book{song2023integrity,
  title={Integrity Constraints on Rich Data Types},
  author={Song, Shaoxu and Chen, Lei},
  year={2023},
  publisher={Springer Nature}
}

@article{al2022data,
  title={Data repair of density-based data cleaning approach using conditional functional dependencies},
  author={Al-Janabi, Samir and Janicki, Ryszard},
  journal={Data Technologies and Applications},
  volume={56},
  number={3},
  pages={429--446},
  year={2022},
  publisher={Emerald Publishing Limited}
}

@article{vaswani2017attention,
  title={Attention is all you need},
  author={Vaswani, Ashish and Shazeer, Noam and Parmar, Niki and Uszkoreit, Jakob and Jones, Llion and Gomez, Aidan N and Kaiser, {\L}ukasz and Polosukhin, Illia},
  journal={Advances in neural information processing systems},
  volume={30},
  year={2017}
}
\end{document}